\documentclass[twocolumn,letterpaper]{IEEEAerospaceCLS}  

\usepackage[]{graphicx}    
\usepackage[colorlinks=true, urlcolor=blue, linkcolor=black, citecolor=black]{hyperref} 
\usepackage{amsmath}
\usepackage{tabularx}
\newcolumntype{C}{>{\centering\arraybackslash}X}
\usepackage{float}
\usepackage{enumitem}
\setlist[itemize]{leftmargin=1.5em}
\newcommand{\ignore}[1]{}  
\begin{document}
\title{Machine Learning Argument of Latitude Error Model for LEO Satellite Orbit and Covariance Correction}

\author{%
Alex Moody\\ 
Draper Scholar\\ 
University of Colorado Boulder\\
Boulder, CO \\
alex.moody@colorado.edu
\and 
Penina Axelrad\\
University of Colorado Boulder\\
Boulder, CO \\
penina.axelrad@colorado.edu
\and
Rebecca Russell\\
The Charles Stark Draper Laboratory, Inc.\\
Cambridge, MA\\
rrussell@draper.com
\thanks{\footnotesize 979-8-3315-7360-7/26/$\$31.00$ \copyright2026 IEEE}              
}

\maketitle

\thispagestyle{plain}
\pagestyle{plain}

\maketitle

\thispagestyle{plain}
\pagestyle{plain}

\begin{abstract}
Low Earth orbit (LEO) satellites are leveraged to support new position, navigation, and timing (PNT) service alternatives to GNSS. These alternatives require accurate propagation of satellite position and velocity with a realistic quantification of uncertainty. It is commonly assumed that the propagated uncertainty distribution is Gaussian; however, the validity of this assumption can be quickly compromised by the mismodeling of atmospheric drag. We develop a machine learning approach that corrects error growth in the argument of latitude for a diverse set of LEO satellites. The improved orbit propagation accuracy extends the applicability of the Gaussian assumption and modeling of the errors with a corrected mean and covariance. We compare the performance of a time-conditioned neural network and a Gaussian Process on datasets computed with an open source orbit propagator and publicly available Vector Covariance Message (VCM) ephemerides. The learned models predict the argument of latitude error as a Gaussian distribution given parameters from a single VCM epoch and reverse propagation errors. We show that this one-dimensional model captures the effect of mismodeled drag, which can be mapped to the Cartesian state space. The correction method only updates information along the dimensions of dominant error growth, while maintaining the physics-based propagation of VCM covariance in the remaining dimensions. We therefore extend the utility of VCM ephemerides to longer time horizons without modifying the functionality of the existing propagator.
\end{abstract} 
\tableofcontents

\section{Introduction}
Positioning, navigation, and timing (PNT) using global navigation satellite systems (GNSS) is susceptible to jamming, spoofing, and signal degradation. These limitations have driven the development of alternative PNT methods leveraging Low Earth Orbit (LEO) satellites that were not designed for PNT. Unlike conventional satellite-based navigation systems, these non-cooperative satellites do not embed ephemerides in their transmitted signal. Users of non-cooperative LEO satellite transmissions must therefore obtain the satellite ephemerides by other means, such as using active radar tracking or passive tracking of the signal transmission from known locations. Regardless of the source, updated ephemerides must be downloaded from the tracking service provider and propagated to the time of measurement. The resulting PNT solution accuracy is therefore driven by the orbit propagation accuracy since the last ephemerides update. Improved propagation accuracy over long time horizons would reduce the required ephemerides update frequency and improve the robustness of the alternative PNT system. 

\begin{figure}
    \centering
    \includegraphics[width=0.95\linewidth]{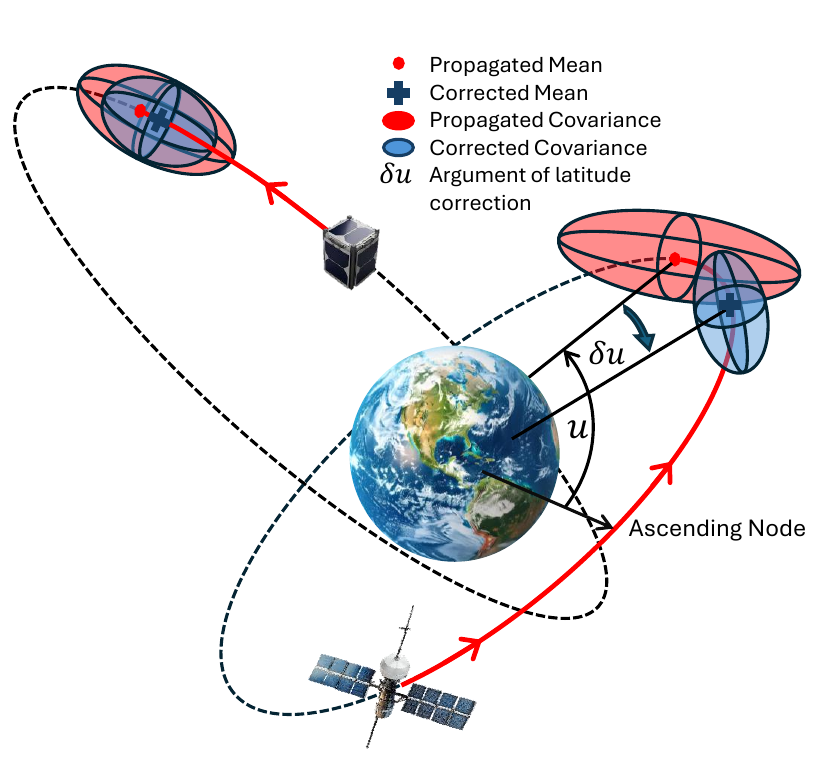}
    \caption{Machine learning model predicts errors and variance in the argument of latitude to correct both the propagated orbit and covariance for a variety of LEO orbit and satellite types.}
    \label{fig:intro-drawing}
\end{figure}
Machine learning (ML) techniques are attractive for improving orbit prediction accuracy because they are well suited to learn the complex patterns in propagation errors given ample historical tracking data \cite{caldas2024machine}. A common hybrid ML approach uses the output of the machine learning model to correct an existing orbit propagator. Peng and Bai have explored using support vector machines, artificial neural networks (ANNs), and Gaussian processes (GPs) to forecast errors in two line element (TLE) ephemerides propagated with the simple general perturbations model 4 (SGP4) propagator \cite{peng2019comparative}. They developed an approach to fuse the propagated solution with the machine learning model output using an Extended Kalman Filter and Particle Filter  \cite{peng2021fusion}\cite{peng2022improving}, assuming both the propagated and correction uncertainty can be trusted. They tested their approach on a real TLE dataset containing 111 satellites in different orbital regimes by training a model for each satellite and error dimension in the local orbit frame defined by the radial, along-track and cross-track axes (RSW) \cite{peng2023medium}. These works demonstrated the practicality and power of the hybrid ML approach, but do not focus on generalization of a single model across different satellite types and orbits. In addition, they do not consider how inaccurately propagated covariance would affect the fusion approach or how to forecast correlations between error dimensions.

Several recent works \cite{kouba2024improved}\cite{huang2025neural} have focused on generalization across the Starlink constellation using the framework developed by Peng and Bai. These works demonstrated improved SGP4 orbit propagation accuracy using a time delay neural network (TDNN) and ANN to predict corrections in all six position and velocity dimensions for 1 hour and 3.5 day propagation windows, respectively. However, these methods do not predict corrections to the propagated covariance. Furthermore, all Starlink satellites have similar physical characteristics and orbit shapes, reducing the generalization complexity.

\begin{figure}
    \centering
    \includegraphics[width=0.9\linewidth]{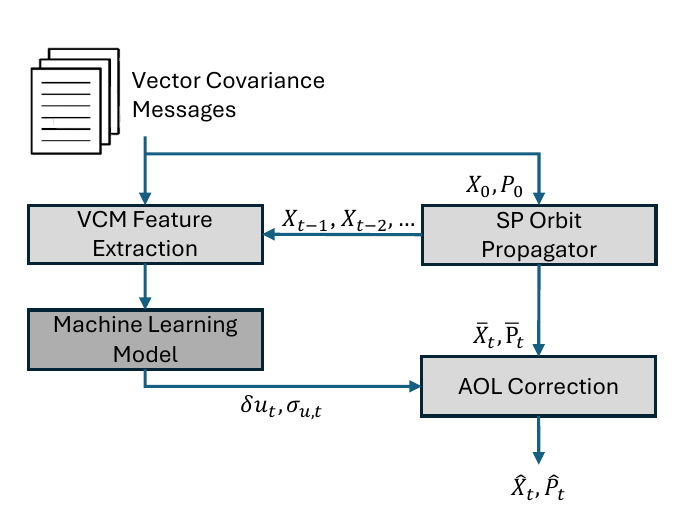}
    \caption{Orbit propagation correction pipeline using VCMs, open source SP propagator, and machine learning model.}
    \label{fig:flowchart}
\end{figure}

We build upon this prior work by expanding the training dataset to include a wide variety of LEO satellites with different physical characteristics, orbit sizes, and orbit shapes. Furthermore, our work is the first to extend the hybrid ML approach to more accurate ephemerides produced by the Space Force \cite{lavezzi_2024_13892132} in vector covariance message (VCM) format \cite{vcm_definition} and a special perturbations (SP) propagator. By extending the hybrid approach to more accurate ephemerides and propagators \cite{conkey2022assessing}, we focus on improving orbit and covariance accuracy over a week-long time horizon. Additionally, we simplify the modeling approach from prior work by only predicting corrections for the dominant error growth in LEO, which accumulates in the argument of latitude (AOL) dimension (as illustrated in Figure \ref{fig:intro-drawing}). Finally, we correct along dimensions with the most severe mismodeling by mapping the one dimensional model outputs to the Cartesian frame. This correction includes the mismodeled correlation terms in the covariance matrix. Figure \ref{fig:flowchart} shows how the VCMs, machine learning models, orbit propagator, and correction approach work together to improve orbit propagation.

Section \ref{sec:dataset_generation} discusses the public VCM dataset and how it is used with the SP propagator to compute propagation errors. Section \ref{sec:aol_parameterization} discusses the mapping of argument of latitude error and variance to the Cartesian frame. Section \ref{sec:feature-extraction} discusses the feature set and ML models used. Finally, Section \ref{sec:results} discusses the performance of the models and correction approach using 1000 LEO satellites over a seven day forecasting window.

\section{Propagation Error Dataset Generation}
\label{sec:dataset_generation}

\subsection{VCM Dataset}
\label{sec:vcm-dataset}

The VCM dataset used in this work was originally created by the United States Space Force Space Command (\mbox{USSPACECOM}) and was distributed publicly by Lavezzi et al, \cite{lavezzi2024early} with permission. Each VCM contains state information for satellites at a single epoch, along with parameters used in the dynamics models for propagation and orbit determination. The original dataset contains 22,303 resident space objects (RSOs) spanning various types of satellites and orbits. Table \ref{tab:VCM-Parameters} describes the VCM fields used in our approach.

\begin{table}[tb]
\renewcommand{\arraystretch}{1.3}
\caption{\textbf{Summary of VCM parameters used in modeling and dataset generation}}
\centering
\begin{tabular}{|>{\centering\arraybackslash}m{3cm}|m{4.5cm}|}
\hline
\bfseries Parameter & \bfseries Description \\
\hline\hline
Epoch time\ (UTC) & Orbital estimate epoch time, expressed in Coordinated Universal Time.\\
\hline
J2K Pos/Vel X-Y-Z\newline ($km$, $km/s$) & The position and velocity vector components of the satellite expressed in Earth-Centered Inertial reference frame.\\
\hline
Ballistic Coef. ($m^2/kg$)& Estimated ballistic coefficient at epoch.\\
\hline
Solar Rad Press Coeff ($m^2/kg$)& Estimated solar radiation pressure coefficient at epoch.\\
\hline
Geo Pot& Degree and order of geopotential model used.\\
\hline
Drag& Atmospheric density model used.\\
\hline
F10, F10A& F10 (10.7 cm) solar flux at epoch time and 81-day average solar flux.\\
\hline
Sigma U-V-W-Ud-Vd-Wd ($km$, $km/s$)& Estimated position and velocity error standard deviations in RSW Frame.\\
\hline
\end{tabular}
\label{tab:VCM-Parameters}
\end{table}

An important limitation of the public dataset is that it only provides the estimation error standard deviations for position and velocity instead of a full covariance matrix. Furthermore, the VCMs truncate most velocity sigmas to zero because they only provide 0.1 m/s precision. Without correlations and valid velocity uncertainty, the propagated covariance is expected to be significantly inaccurate. Nevertheless, the position sigmas provide an approximate initial uncertainty that can be used for a preliminary analysis of our covariance correction approach. 

We reduced the full dataset to a list of 2,486 LEO satellites using information from Space Track. The list of all LEO satellites currently in orbit\footnote{\nolinkurl{https://www.space-track.org/basicspacedata/query/class/satcat/
PERIOD/<128/DECAY/null-val/CURRENT/Y/format/csv}} was first downloaded from Space Track and then filtered to those satisfying the following criteria:
\begin{itemize}
    \item Classified as having a large radar cross section (RCS) by Space Track
    \item Perigee altitude less than 1200 km
    \item Not classified as "Debris''
    \item Not Starlink or OneWeb
\end{itemize}

\begin{figure}
    \centering
    \includegraphics[width=1\linewidth]{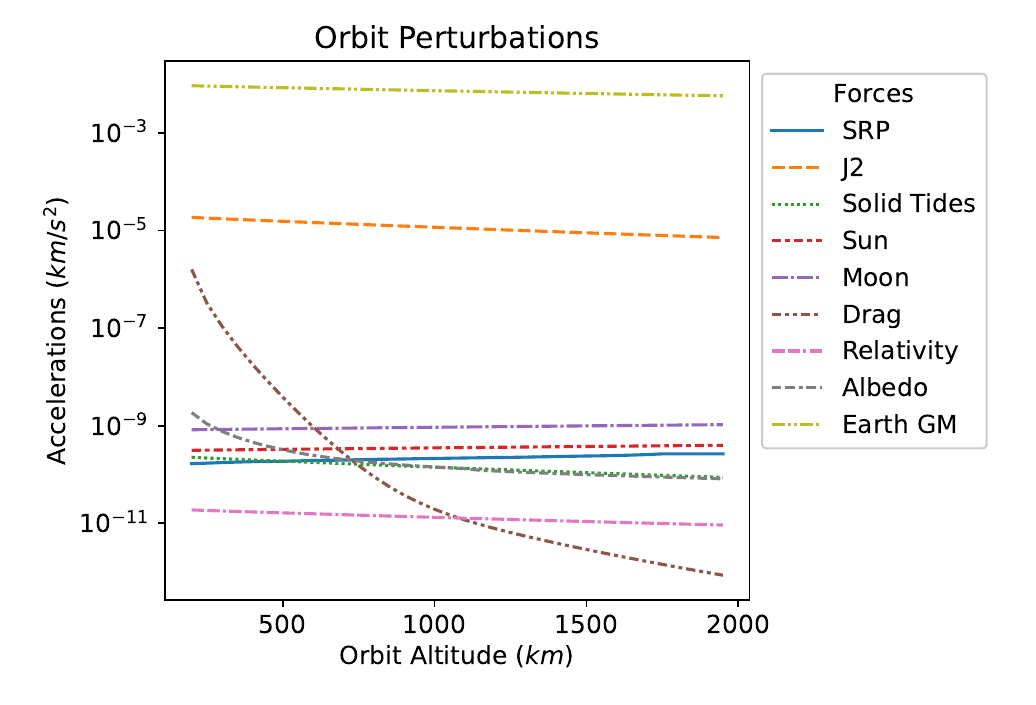}
    \caption{Accelerations due to individual force models in Orekit orbit propagator over various altitudes for an example satellite in an initially circular orbit with a ballistic coefficient of 0.13 $m^2/kg$.}
    \label{fig:perturbations_plot}
\end{figure}

Large RCS satellites are chosen because they are easier to track with ground stations. We also set a perigee altitude threshold of 1200 km to focus on satellites most affected by mismodeled atmospheric drag. Figure \ref{fig:perturbations_plot} shows how the magnitude of the atmospheric drag perturbation increases rapidly as the orbital altitude decreases below 1200 km. We also remove the maneuvering OneWeb and Starlink constellations in this study to explore generalization across satellites with natural dynamics. Although other active payloads besides Oneweb and Starlink maneuver, removing the large constellations helps increase the percentage of non-maneuvering satellites in the dataset. This list is reduced further to 1,000 objects by removing the remaining maneuvering satellites.

\subsection{Propagator Setup}
\label{sec:prop_setup}

The Orekit \cite{maisonobe_2024_13950582} open source orbit modeling software is used to propagate the VCM states and covariances. The propagator is configured to closely match the dynamic models in the VCM files. The force models shown in Table \ref{tab:force_models} are integrated with Dormand-Prince 8(5,3) \cite{maisonobe_2023_8418401} to propagate each set of VCM initial conditions. By using a high order gravity model and including other significant perturbations in LEO orbit, errors in drag modeling are left as the most significant error source in the propagator.    

\begin{table}
\caption{Orekit Force Models}
\centering
\renewcommand{\arraystretch}{1.3}
\begin{tabular}{ |c| }
    \hline
    \textbf{Perturbation Force Models} \\
    \hline\hline
   Solar Radiation Pressure \\
   70 $\times$ 70 Gravity (EGM-96) \\
   Solid Earth Tides \\
   Sun Third Body \\
   Moon Third Body \\
   Atmospheric Drag (NRLMSISE-00) \\
   Relativity \\
   Albedo (Knocke) \\
   \hline
\end{tabular}
\label{tab:force_models}
\end{table}
\normalsize

The ballistic coefficients $B_C$ provided in the VCMs are used to initialize the drag model for each satellite. Orekit has separate parameters for the coefficient of drag, cross sectional area, and mass. The ballistic coefficient is split up into these parameters by fixing the satellite mass $m$ and adjusting the cross sectional area $A$ to be consistent with the ballistic coefficient $B_C$ provided in each VCM epoch:
\begin{equation}
    A = \frac{mB_c}{C_d}
\end{equation}
where $C_d$ is fixed at 2.2.  This method was compared with using masses from the DISCOS \cite{Klinkrad1991DISCOSE} dataset and similar propagation performance was observed. 

The propagator state is initialized using the position and velocity provided for a VCM epoch. Given the velocity sigma values in the VCMs are generally truncated to zero, we set them to the largest value that would not be captured using the file precision. The initial covariance is therefore likely to be somewhat conservative, given the inflation of the velocity sigmas and assumption of no correlation between states in the RSW frame. 

\subsection{Propagation Error Computation}
The propagation errors are computed by initializing the propagator at one VCM epoch and propagating forward to future VCM epochs. The VCM states and propagated states are then converted to osculating orbital elements from Earth Centered Inertial (ECI) positions and velocities. The argument of latitude errors are computed by differencing the osculating orbital elements at each VCM epoch in a 7 day window. The VCM epochs are asynchronous across satellites, so the computed errors do not have common timestamps or frequency. 

Although this error computation ignores noise in the reference states, it is a valid approximation, given that the VCM uncertainties are small relative to the systematic orbit errors we observe after 1.5 days of propagation. The median norm position standard deviation reported in the VCMs used for this study is 14.7 meters, which is significantly less than the 134.5 meter median norm propagation error between 1-1.5 days. 

\section{Argument of Latitude Error Modeling}
\label{sec:aol_parameterization}
Although LEO SP propagators can leverage more accurate force models than SGP4, they are still limited by mismodeled atmospheric drag \cite{ray2022framework}. Drag force mismodeling is caused by errors in the satellite specific ballistic coefficient and atmospheric density models. Operational semi-empirical atmospheric density models such as NRLMSISE-00 and JB2008 fit an analytical model to historical calibration data to forecast atmospheric densities \cite{caldas2024machine}. These models are especially inaccurate at forecasting density during periods of high solar activity \cite{brandt2020simple}. The drag force primarily acts opposite to the direction of spacecraft motion, causing errors to accumulate in the argument of latitude when expressed in orbital elements. By modeling errors in the argument of latitude, we capture the dominant error growth in position and velocity with a single dimension.

\subsection{State Transformation}

Hayek and Kassas provide a convenient transform of errors in the argument of latitude to the RSW frame \cite{hayek2024analysis}. They only consider the argument of latitude error because other orbital elements will accumulate errors at a much slower rate. Therefore, propagated orbital elements are used to map the error in argument of latitude ($\delta u$) to RSW errors. Since the argument of latitude defines a spacecraft position along an orbit, it can only be used to capture errors in the orbit plane. Equation \ref{eq:aol_parameterization_position} shows no errors are mapped to the orbit normal direction for both position and velocity. 

\begin{align}
\delta &X_{RSW} = \begin{bmatrix} 
    r(1-\cos{\delta u})\\
    \frac{r\{2\sin{\delta u}-e[\sin(\omega-u-2\delta u))+\sin(u-\omega)]\}}{2(e\cos{(u-\omega)}+1)}\\
    0 
    \end{bmatrix} \label{eq:aol_parameterization_position}\\
\nonumber\\
\delta &\dot{X}_{RSW} = \nonumber\\
& \begin{bmatrix} 
-\frac{\mu}{h}\sin(\delta u) \\
\frac{\mu\{-2\cos{\delta u}+2+e[-\cos(\omega-u-2\delta u)+2\cos(\delta u+u-\omega)-cos(u-\omega)]\}}{2h(e\cos{(u-\omega)}+1)} \\
0 
\end{bmatrix}\label{eq:aol_parameterization_velocity}
\end{align}

where $\mu$ is the gravitational parameter for Earth and $e,u,\omega,f,h$ are the osculating eccentricity, argument of latitude, argument of perigee, true anomaly and angular momentum for the propagated state.
The argument of latitude correction error uncertainty is mapped to the RSW and ECI coordinate frames with
\begin{align}
\label{eq:aolcov_transform}
\Lambda &\overset{\Delta}{=} \begin{bmatrix}
    \frac{\partial \delta X_{RSW}}{\partial\delta u}\nonumber \\
    \frac{\partial \delta \dot{X}_{RSW}}{\partial\delta u}\nonumber
\end{bmatrix}\\
&= \begin{bmatrix}
    r\sin(\delta u)\\\\
    \frac{r\{\cos{\delta u}+e\cos(\omega-u-2\delta u))\}}{(e\cos{f}+1)}\\\\
    0\\\\
    -\frac{\mu}{h}\cos(\delta u)\\\\
    \frac{\mu\{\sin{\delta u}-e[\sin(\omega-u-2\delta u)+sin(\delta u + u -\omega)]\}}{h(e\cos{f}+1)}\\\\
    0\\\\
\end{bmatrix}
\end{align}

\begin{align}
    P_{ECI} &= R_{ECI}^{{RSW}\ T}\Lambda P_{u}\Lambda^TR_{ECI}^{RSW}
\end{align}

where $P_{u}$ is the argument of latitude covariance, $R_{ECI}^{RSW}$ is the rotation matrix from ECI to RSW and $P_{ECI}$ is the argument of latitude variance expressed in the ECI frame.

\subsection{Correction Approach}

\begin{figure*}[htbp]
    \centering
    \includegraphics[width=1\linewidth]{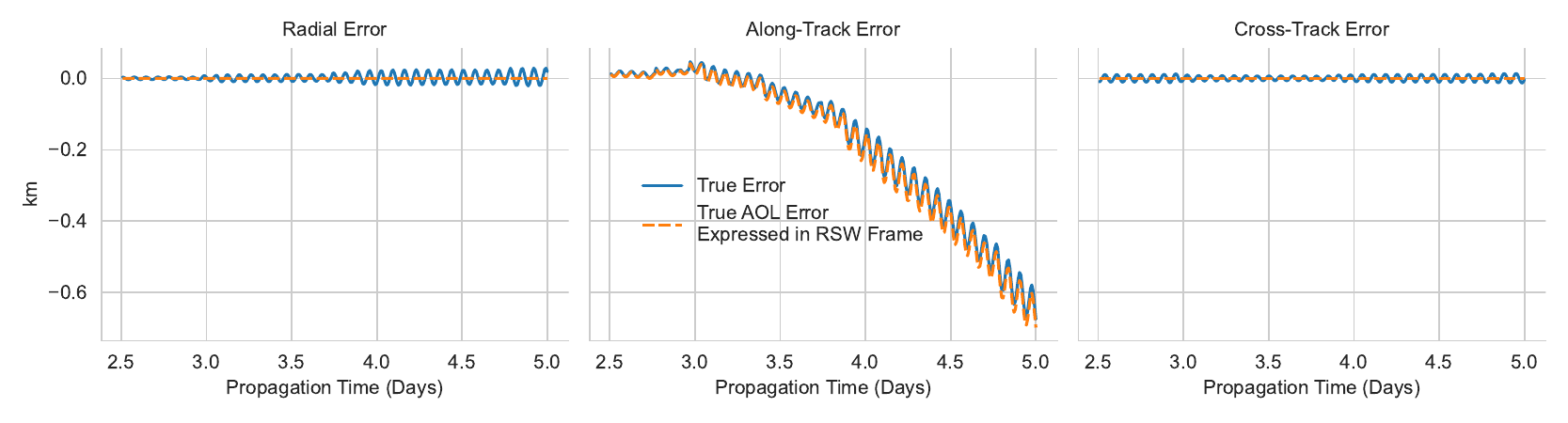}
    \label{fig:position_aol_mapping}
    \includegraphics[width=1\linewidth]{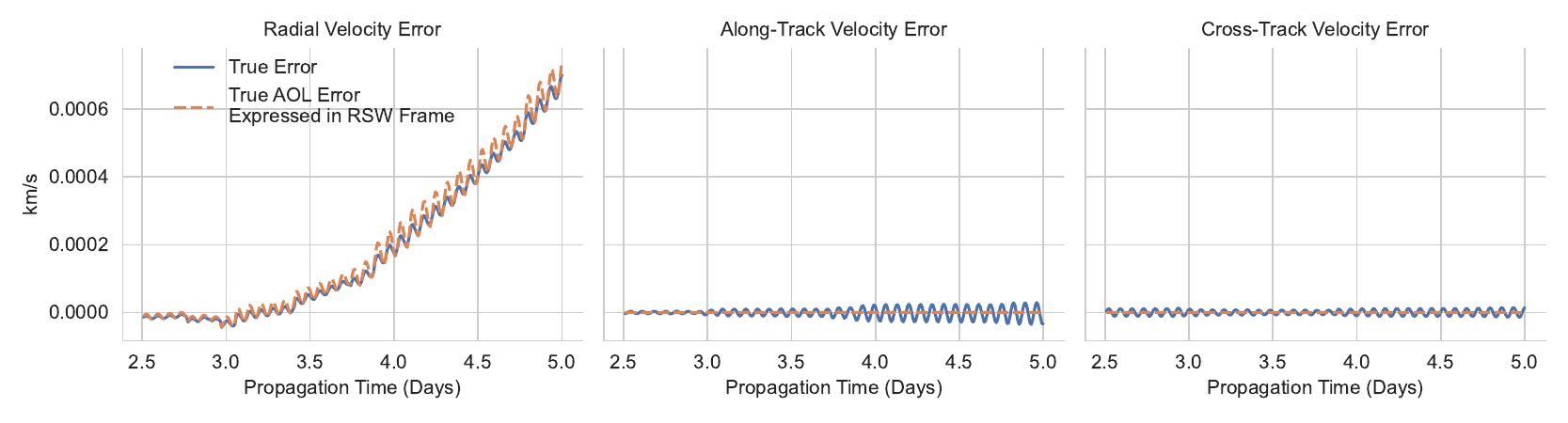}
    \label{fig:velocity_aol_mapping}

    \caption{Example of error growth for propagation of a single VCM (NORAD ID 47). True AOL error maps to predominantly along-track position and radial velocity dimensions. Errors in the remaining dimensions are caused by errors in other orbital elements and therefore are not captured by argument of latitude. } 
    \label{fig:aol_mapping_rsw}
\end{figure*}

\begin{figure}[htbp]
    \centering
    \includegraphics[width=2.5\linewidth]{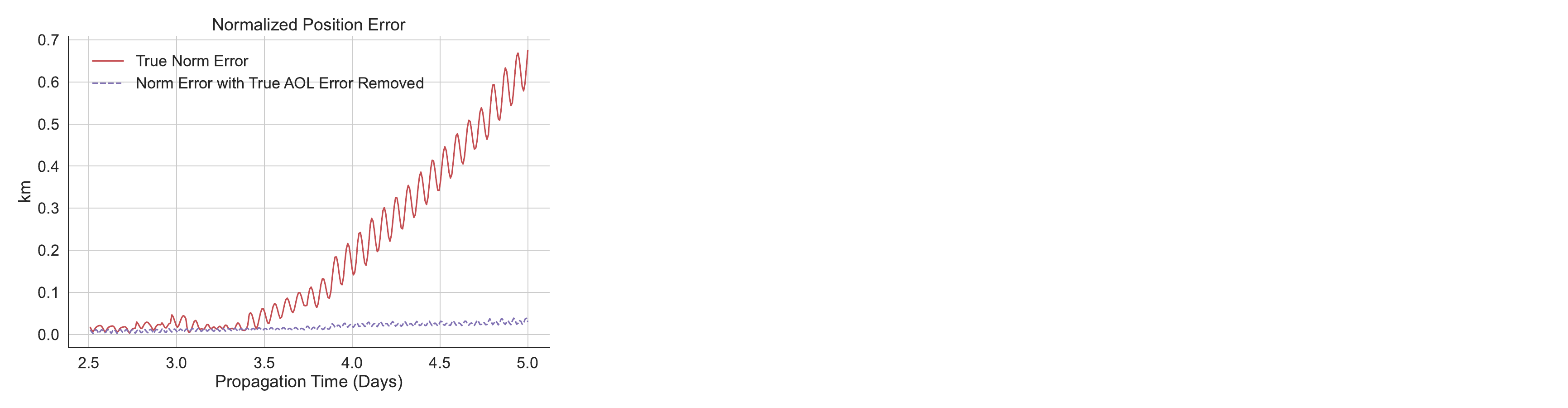}
    \caption{Over 95\% of the norm position error is captured by the argument of latitude after 5 day propagation (NORAD ID 47).}
    \label{fig:aol_correction_norm_error}
\end{figure}

Our approach corrects the dimensions most affected by drag mismodeling while maintaining accurate propagation information in the remaining dimensions. Figure \ref{fig:aol_mapping_rsw} shows how the true argument of latitude error mapped to the RSW frame captures the secular drift in along-track position and radial velocity. Although the argument of latitude error also maps to the radial position and along-track velocity, it is not responsible for the dominant error growth in those dimensions. Errors in other orbital elements create zero mean oscillations with slowly growing amplitudes. In Figure \ref{fig:aol_correction_norm_error} we show how subtracting the mapped true argument of latitude error from the propagated states reduces most of the error for a LEO satellite. For instance, by removing the drift in the along-track dimension, we reduce the norm position error by over 95\%. We therefore only correct the marginal distribution of the along-track position and radial velocity variables.


We modify the fusion approach presented in \cite{peng2021fusion} to account for scenarios where the propagated covariance is inaccurate. When the propagator does not properly account for process noise, the propagated covariance becomes unrealistic, and typically overconfident. If the propagated covariance is overconfident and a Kalman update is performed with the ML output, the propagated state would be weighted heavier than the ML correction. We instead inflate the covariance in the along-track position and radial velocity dimensions before performing the Kalman update. This allows for the ML model to inflate the covariance if necessary. Consequently, the ML model must output variances consistent with its error predictions for this method to be robust.   

The propagated state and covariance in ECI are defined as
\begin{align}
    X_{k|k-1},\  P_{k|k-1}
\end{align}
where $X_{k|k-1}$ is a $6\times 1$ matrix and $P_{k|k-1}$ is a $6 \times 6$ matrix.
The one dimensional predicted argument of latitude error and uncertainty are
\begin{align}
    \delta u_k,\ P_{u,k}
\end{align}

The propagated state is corrected with the ML predicted error to get a new state estimate ${\hat{X}}_k$. $P_{k|k-1}$ is inflated along the correction axes with a scaled version of the ML predicted covariance $P_{RSW,k}$.

\begin{align}
    &\delta u \rightarrow \delta X_{RSW}, \delta\dot{X}_{RSW}\text{ (computed with (\ref{eq:aol_parameterization_position}) \& (\ref{eq:aol_parameterization_velocity}))}\nonumber\\
    &\hat{X}_{k|k} = X_{k|k-1} + R_{ECI}^{{RSW}\ T}
        \begin{bmatrix}
            \delta X_{RSW}\\
            \delta \dot{X}_{RSW}
        \end{bmatrix}_k\\
    & P_{RSW,k} = \Lambda P_{u}\Lambda^T + P_{RSW,0}\\  
    & Q  = \alpha\  \text{diag}(0,1,0,1,0,0) P_{RSW,k}\ \text{diag}(0,1,0,1,0,0)\\
    & P_{k|k-1} = P_{k|k-1} + Q
\end{align}

where $\alpha$ is a scale factor we set to 1e6 and the diagonal matrices are used to select the marginal distribution we want to update. The nonzero terms of the process noise covariance matrix $Q$ are therefore much larger than the unscaled ML predicted covariance. This makes it possible for the ML covariance update to inflate the propagated covariance. Furthermore, the initial VCM covariance is combined with the predicted covariance because mapping one dimensional variance into multiple dimensions through a linear transformation creates a singular matrix. Adding the initial covariance makes $P_{RSW,k} $ positive definite but should not significantly affect the corrected uncertainty for long propagation times where $P_{k|k-1} >>P_{RSW,0}$. We finally perform a Kalman update with the ML predicted covariance.

\begin{align}
    & H =
        \begin{bmatrix}
            0&1&0&0&0&0\\
            0&0&0&1&0&0
        \end{bmatrix}R_{ECI}^{{RSW}}
\end{align}
{\normalsize
\begin{align}
    & P_{r\dot{s},k} = \begin{bmatrix}
        P_{RSW,k}(2,2)&P_{RSW,k}(2,4)\\
        P_{RSW,k}(4,2)&P_{RSW,k}(4,4)
    \end{bmatrix}\\
    &K = P_{k|k-1}H^T(HP_{k|k-1}H^T+P_{r\dot{s},k})^{-1}\\
    &\hat{P}_{k|k} = (I-KH)P_{k|k-1}(I-KH)^T+KP_{r\dot{s},k}K^T
\end{align}}

where $P_{r\dot{s},k}$ is the $2\times2$ marginal distribution of $P_{RSW,k}$ in the along-track position and radial velocity. The measurement matrix H maps the propagated covariance to this two dimensional frame. We therefore correct the mismodeled components of the distribution without corrupting accurately propagated information in other dimensions.

\section{Machine Learning Approach}
We considered two machine learning models for forecasting argument of latitude errors. The first is a time-conditioned feed forward neural network (TCNN) \cite{bishop:2006:PRML}, and the second is a heteroscedastic Gaussian process (HGP) \cite{kersting2007most}. These two models predict means and covariance values, allowing us to compare the performance of non-parametric and parametric models for this regression problem.

\subsection{Feature Extraction}
\label{sec:feature-extraction}
The selection of features was an iterative process through testing on smaller datasets and monitoring model performance. Orbital elements used as features were derived from VCM positions and velocities. The features used in the NN are as follows:
\begin{itemize}
    \item Reverse propagation times and along-track errors up to 2 days in the past
    \item Altitude at perigee
    \item Eccentricity 
    \item $\cos(f)$
    \item $\cos(i)$
    \item $B_C$ (VCM Parameter)
    \item $F_{10.7A}$ (VCM Parameter)
    \item Payload Classification (one-hot encoding)
    \item Rocket Body Classification (one-hot encoding)
    \item $\Delta t_{prop}$
\end{itemize}

where $f$ is the propagated true anomaly and $i$ is the propagated inclination. The cosine is taken of true anomaly to encode distance from perigee or apogee. For eccentric orbits, the propagation error will expand around apogee and shrink around perigee due to orbital dynamics. The cosine is also taken of inclination to map the angle to a value between -1 and 1.

Since the VCM epoch spacing differs between satellites, a maximum of 11 epochs in the reverse direction are used to generate features. Satellites that do not have 11 epochs within the 2 day window are padded with reverse propagation errors of zero at time zero. Additionally, the payload and rocket body classifications are pulled from DISCOS and are binary values to encode the satellite type. The 11 reverse propagation errors, 11 reverse propagation times, and the 9 other variables add up to a feature vector with 31 dimensions.

Although HGPs are technically non-parametric, they still require optimization of the kernel function parameters that define the covariance functions between data points. The training of a GP has computational complexity of $\mathcal{O}(n^3)$ and prediction has computational complexity of $\mathcal{O}(n^2)$ \cite{van2024scalable}. To reduce the computational load of the HGP, only the longest reverse propagation epoch is used as a feature, along with altitude at perigee, $B_c$ and $\Delta t_{prop}$. We arrived at this subset of features through multiple rounds of training and testing the HGP. 

\subsection{Time-Conditioned Feed Forward Neural Network}
The TCNN consists of an input layer, two hidden layers and one output layer. The output layer is a univariate Gaussian characterized by the output mean and variance. Each hidden layer has 128 dimensions and the input layer has 31 dimensions. We train the TCNN on the argument of latitude propagation error using Gaussian maximum likelihood.

The predicted argument of latitude error is assumed to follow a normal distribution $\delta u \sim \mathcal{N}(\mu_{\delta u},P_u)$. The probability density function for the univariate Gaussian is
\begin{align}
    p(\delta u|z)=\frac{1}{\sqrt{2\pi\sigma(z)^2}}\exp\left[-\frac{(\delta u - f(z))^2}{2\sigma(z)^2}\right]
\end{align}
where $f(z)$ is the learned mean function and $\sigma(z)$ is the learned variance function. We define our loss function $\mathcal{L}$ as the negative logarithm of the univariate Gaussian likelihood. We train the model by minimizing the loss function through gradient decent and optimize $f$ and $\sigma$ simultaneously. 
\begin{align}
    \mathcal{L} = \frac{1}{2}\left[\frac{(\delta u-f(z))^2}{\sigma(z)^2} + \ln\sigma(z)^2\right]
\end{align}
We use an exponential activation function for the output variance to stabilize training \cite{russell2021multivariate}.

\subsection{Heteroscedastic Gaussian Process}

The general expression for a GP is given by Rasmussen \cite{Rasmussen2006Gaussian} as,
\begin{equation}
    f(x) \sim \mathcal{GP}(m(x),k(x,x'))
\end{equation}
where we define the mean function $m(x)$ as zero \cite{Rasmussen2006Gaussian}. The kernel function $k(x,x')$ defines the covariance between two function outputs. Basis functions are typically chosen to produce higher correlation values for inputs that are closer together. We use the common Gaussian basis function.
\begin{equation}
    k(x_i,x_j)=\exp\left(-\frac{(x_i - x_j)^2}{\theta}\right)
\end{equation}
where $\theta$ is the length scale for the basis functions. The covariance matrix for the GP consists of an $N\times N$ matrix where each element $(i,j)$ contains the output of the kernel function for the inputs $x_i,x_j$. For a GP with zero mean, the covariance matrix fully defines the joint distribution of all function outputs.

Given a set of observations $Y$ and features $X$, the predictive distribution can be found by first estimating the kernel function parameters. They can be found by maximizing the log likelihood function $\log{p(Y|X,\theta)}$. Once the parameters are determined, test features $x_*$ can be added to the GP covariance matrix using the kernel function to correlate $x_*$ with all other inputs. The predictive distribution is therefore given by the conditional probability of $y_*$ given $Y$. Since all of the function outputs are jointly Gaussian, the conditional distribution will also be Gaussian and is given by equation \ref{GP_predictive_dist} and \ref{GP_predictive_dist2}
\begin{equation}
    \label{GP_predictive_dist}
    \mu_* = k(x_*,X)[K+\beta^{-1}I]^{-1}Y
\end{equation}
where $k(x_*,X)$ is the covariance function evaluated with $x_*$ and all other inputs in training set $X$. $\beta$ is the precision term representing the constant noise on the observations.
\begin{equation}
    \label{GP_predictive_dist2}
    \sigma^2_* = k(x_*,x_*)-k(x_*,X)[K+\beta^{-1}I]^{-1}k(X,x_*)
\end{equation}
The advantage of GP prediction is that it naturally accounts for epistemic uncertainty. If an input feature vector is far from all of the feature vectors in the model, then the GP would predict an output distribution with large variance.

\begin{figure}
    \centering
    \includegraphics[width=0.85\linewidth]{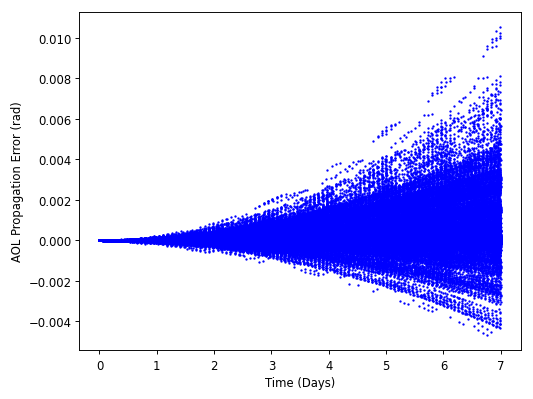}
    \caption{AOL errors used to train the ML models consisting of 50 VCMs propagated 7 days for 800 satellites.}
    \label{fig:AOL_error_example}
\end{figure}

We also recognize that random modeling errors accumulate during propagation and create noise distributions as a function of the propagation time. We show an example of the varying noise profile for argument of latitude errors in Figure \ref{fig:AOL_error_example}. We therefore use a heteroscedastic Gaussian Process to model the noise with a latent GP. 

We use the \textit{hetGPy} python package for an optimized heteroscedastic GP framework \cite{o2025hetgpy}. The package extends the \textit{hetGP} R package that is designed to handle large datasets through efficient use of replicated inputs \cite{binois2021hetgp}. Although the input data is unlikely to have many exact replicates due to continuous scales on most input features, the package is efficient at optimizing the hyperparameters for large datasets. Even with the optimized software, training the model with the full dataset and feature list is prohibitively slow. We discuss the compromises made during GP training in the following section.

\section{Experimental Results}
\label{sec:results}
The machine learning models are trained on data from 800 satellites randomly picked from the 1000 satellite dataset. They are tested on data from the remaining 200 satellites covering the same time window. This split of the training and validation data allows for testing of generalization across satellite orbits and physical characteristics while ignoring temporal effects on orbit propagation. Generalization to future time windows is briefly investigated; however, it will be the focus of future work.

We test each model by predicting argument of latitude errors for validation feature vectors. The prediction error is the difference between the model output and the validation target errors. The prediction error is then compared with the model predicted covariance. The model prediction means and variances are mapped to the Cartesian RSW frame where the correction approach described in section \ref{sec:aol_parameterization} is applied. We then present the improvements made by the model corrections to the true error distribution and covariance realism.

All features and labels except for the satellite type classifications are normalized with the mean and standard deviation of the training dataset.
\begin{equation}
    x_{normalized} = \frac{x-\mu_x}{\sigma_x}
\end{equation}
where $x$ is a feature variable, $\mu_x$ is the mean of $x$ in the training data, and $\sigma_x$ is the standard deviation of $x$ in training data. During prediction, input features and output error predictions are normalized using the $\mu_x$ and $\sigma_x$ computed with the training set. 

Table \ref{tab:uncorrected-dimension-errors} shows the uncorrected error standard deviations for the training dataset. As expected, most of the error is in the along track position and radial velocity dimensions. The error in the remaining four dimensions is at least two orders of magnitude smaller.
\begin{table}
    \caption{Validation dataset standard deviations in RSW frame after seven days of propagation}
    \renewcommand{\arraystretch}{1.3}
    \centering
    \begin{tabularx}{0.8\columnwidth}{|C|C|C|C|C|C|}
        \hline
         $\sigma_{R_{r}}$ km & $\sigma_{R_{s}}$ km &  $\sigma_{R_{w}}$ km & $\sigma_{V_{r}}$ m/s & $\sigma_{V_{s}}$ m/s & $\sigma_{v_{w}}$ m/s \\
         \hline\hline
         0.23 &12 &0.013 &13 & 0.043 & 0.017 \\
         \hline
    \end{tabularx}
    \label{tab:uncorrected-dimension-errors}
\end{table}

\subsection{Neural-Network Performance}

The TCNN was trained using an Adam optimizer \cite{Kingma2014AdamAM} with a learning rate of $1e^{-3}$ and a batch size of 50,000. 

\begin{figure}
    \centering
    \includegraphics[width=0.9\linewidth]{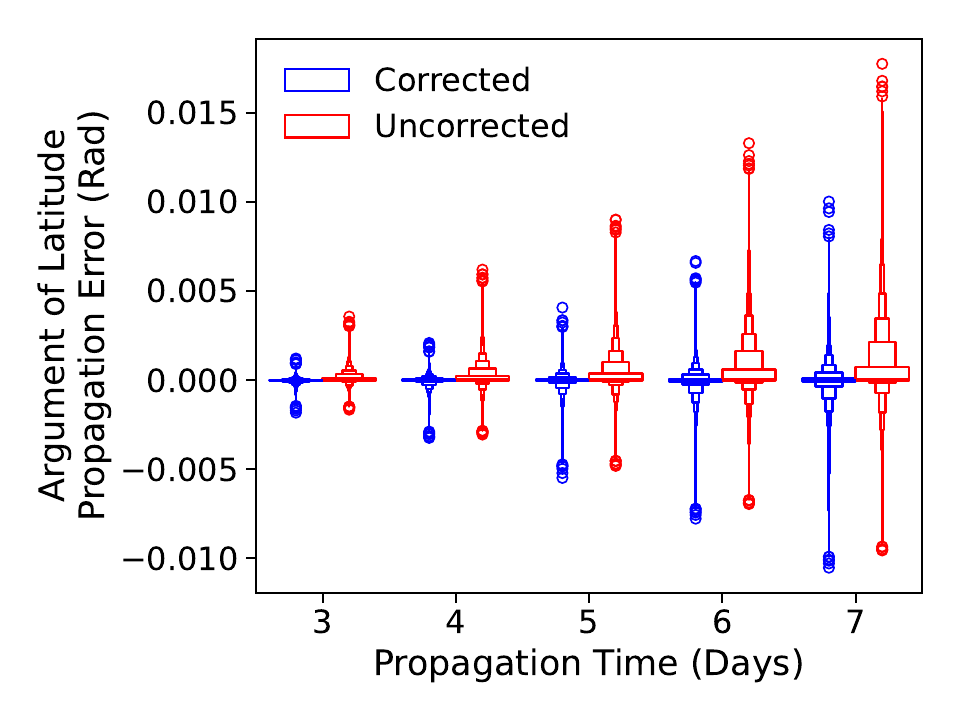}
    \caption{TCNN - Corrected and Uncorrected Error Distributions Over Time.}
    \label{fig:error_time_series} 
\end{figure}

\begin{figure*}
    \centering
    \includegraphics[width=6in]{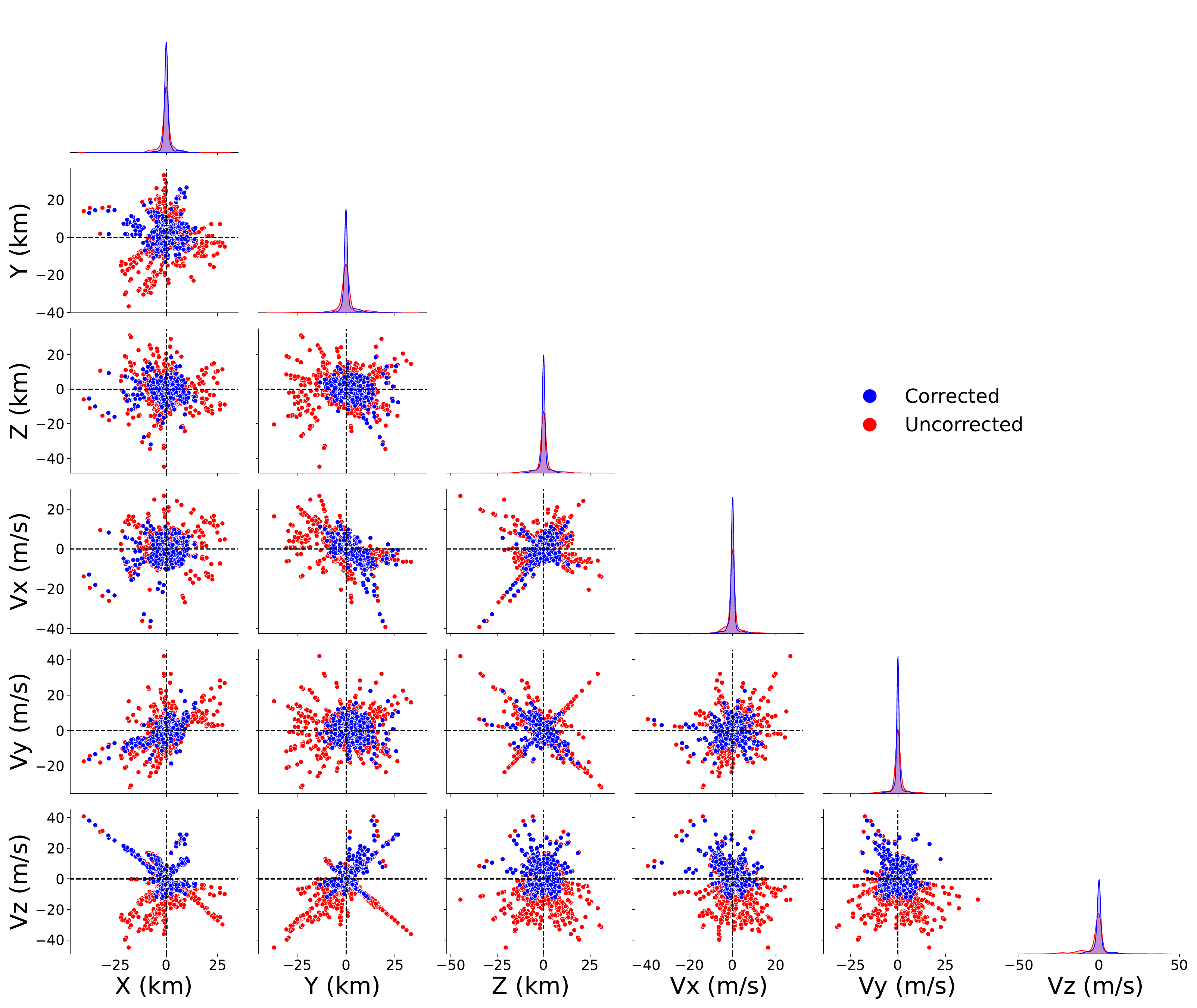}
    \caption{TCNN - Error reduction in all dimensions of ECI Cartesian frame after AOL correction.}
    \label{fig:corner_plot_alcov} 
\end{figure*}

We first compare the corrected AOL error distribution to the uncorrected AOL distribution for each day of propagation in Figure \ref{fig:error_time_series}. Unlike a traditional box plot that only shows four quantiles, the enhanced box plot displays quantiles about the mean decreasing by a factor of two \cite{Waskom2021}. For instance, the innermost boxes represent the 25 and 75 percentiles followed by a set of boxes drawn at the 12.5 and 87.5 percentiles. The extreme points are defined as 3/2 times less than than the lower quantile or 3/2 times greater than the upper quantile. The enhanced box plot shows how the uncorrected true error distribution is skewed in the positive direction and contains extremely long tails. The ML corrections not only improve the skewness but reduce the long tails and spread of the distribution.

Figure \ref{fig:corner_plot_alcov} shows a corner plot of the uncorrected and corrected error distributions in the Earth Centered Inertial (ECI) Cartesian frame. Each histogram along the diagonal represents the error distribution in one of the six error vector dimensions. The two dimensional marginal distributions for all combinations of variables are shown in scatter plots. All of the scatter plots and histograms in the same column have the same x values and all scatter plots in the same row have the same y values. The histograms also match the y values for the scatter plots sharing the same row.

We see here how the one dimensional argument of latitude correction maps to the six dimensional ECI Cartesian state space. The uncorrected distribution is non-Gaussian in many marginal distributions. The histograms show how the spread of the errors decreases in every dimension with the orbit corrections, resulting in tighter and more Gaussian marginal distributions. 

\subsection{Gaussian Process Performance}
In addition to decreasing the number of features as discussed in section \ref{sec:feature-extraction}, we retain every 1,000th sample from the dataset to reduce the memory requirements of the model. This down selection process ensured samples from each satellite remained in the training set. However, it did not ensure that there were samples from every satellite every day. We chose to downsample the dataset by the factor of 1,000  after a few iterations showed it had the best balance of computational resources and performance.

\begin{table*}[htb!]
    \caption{Correction performance for HGP and TCNN for 7 day propagation}
    \renewcommand{\arraystretch}{1.3}
    \centering
    \begin{tabularx}{0.8\textwidth}{|C|C|C|C|C|C|C|}
        \hline
         {Dataset} & $\sigma_{R_{s}}$ km &  $\sigma_{V_{r}}$ m/s & $\sigma_{\|R\|}$ km & $\sigma_{\|V\|}$ m/s &\%{$D_{M_u,C}$} & \%$D_{M_{[R,V]},C}$ \\
         \hline\hline
         Uncorrected & 12 &	13 & 11 & 12 & - & 7.7\\
         \hline
         TCNN & 6.2	& 6.8	&5.5& 6.1 &95&	81\\
         \hline
         HGP & 6.8 &	7.5&	6.1& 6.7 &	97&	81\\
         \hline
    \end{tabularx}
    \label{tab:nominal-stats}
\end{table*}

Table \ref{tab:nominal-stats} shows the performance of the HGP compared with the TCNN and uncorrected distributions; where $\sigma_{R_s}$ and $\sigma_{V_r}$ are the standard deviation of the along-track position errors and radial velocity errors, respectively; $\sigma_{\|R\|}$ and $\sigma_{\|V\|}$ are the standard deviation of the norm position and velocity errors, respectively; $\%D_{M_u,C}$ is the percent of Mahalanobis distance samples that are less than a consistency threshold in the argument of latitude dimension; and $\%D_{M_{[R,V]},C}$  is the same metric but for position and velocity in ECI. The consistency threshold for $\%D_{M_u,C}$ and $\%D_{M_{[R,V]},C}$ is set at the 99th percentile of the $\chi^2$ distribution for one and six degrees of freedom, respectively. It follows that if the reported consistency metric is less than 99\%, there are more distances above the threshold than in the theoretical distribution. The covariance would therefore be overconfident more often than expected. If the reported consistency metric is over 99\%, the covariance would be underconfident more often than expected.

The first line of Table \ref{tab:nominal-stats} shows the uncorrected values. We compute $\%D_{M_{[R,V]},C}$ with the uncorrected errors and covariances to show the consistency of the propagator output. Only 7.7\% of the uncorrected Mahalanobis distance samples are under the consistency threshold. This is not surprising given that process noise is not included and the limited initial covariance information available in the public VCM dataset.

The final two lines in Table \ref{tab:nominal-stats} show the same metrics after correcting the errors and covariances with the model outputs. The corrections from both machine learning models reduce the error sigmas by approximately a factor of two, while increasing the consistency metric from 7.7\% to over 80\%. Although this suggests that the corrected covariance is still somewhat overconfident, the correction significantly improves the covariance consistency. Furthermore, the consistency of the one dimensional model outputs are very close to matching the theoretical distribution. Both models have over 95\% of the argument of latitude Mahalanobis distance samples within the consistency threshold. The variance output by the models is therefore mostly consistent with the true argument of latitude errors, but the mapping to ECI position and velocity cannot correct for all of the inconsistencies in the propagated covariance. 

\begin{figure}
    \centering
    \includegraphics[width=1\linewidth]{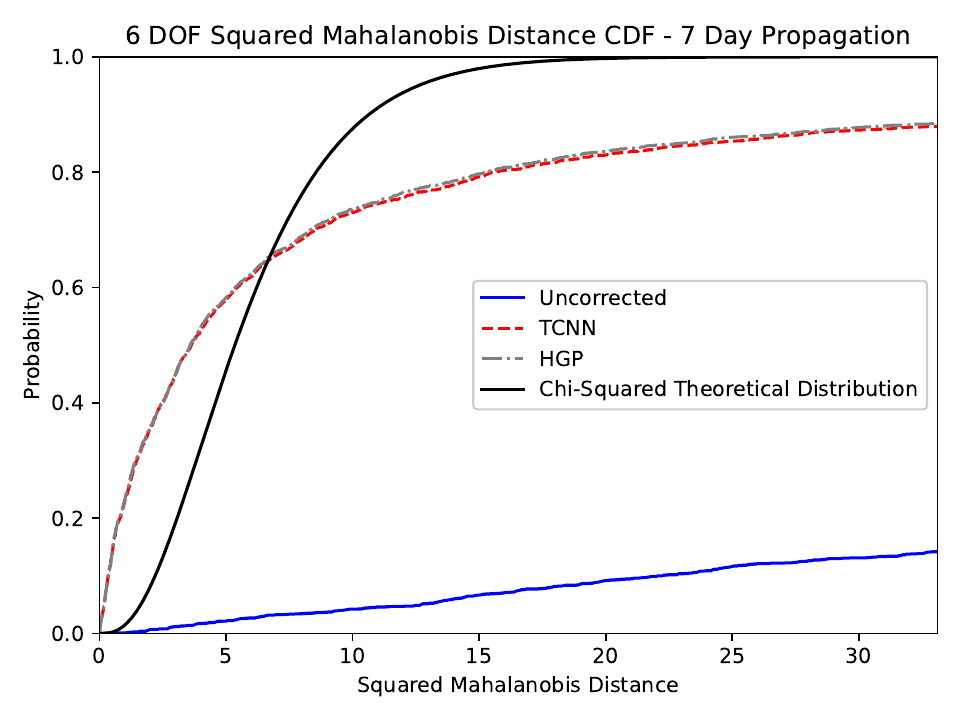}
    \caption{Covariance realism improvement after AOL correction in ECI Cartesian frame.}
    \label{fig:consistency_plot} 
\end{figure}

Figure \ref{fig:consistency_plot} shows the Mahalanobis distance empirical CDF for the uncorrected, TCNN and HGP results in addition to the theoretical chi-squared distribution for six degrees of freedom. The empirical Mahalanobis distance distribution should match the shape of the theoretical distribution if the errors are normally distributed and the variance is scaled correctly. We see that the uncorrected CDF does not follow the theoretical distribution and contains many samples with large Mahalanobis distances. The corrected CDFs using the TCNN and HGP have similar performance and are closer to following the theoretical distribution.

\subsection{Validation using Future Epochs}
We also investigate the generalization of the HGP and TCNN models over time and satellites. We tested time generalization with a validation dataset that contains the same satellites as the original validation dataset, but uses future VCMs to generate the errors. The training dataset therefore does not contain any of the satellites in the validation dataset nor any data for the same time window as the validation dataset. Table \ref{tab:time-generalization} shows how the two models generalize over time. The models still improve the orbit accuracy, but by less than 2 km in position and 2 m/s in velocity. Additionally, the consistency of the TCNN predicted variance is 63.9\%, versus 95.3\% without time generalization. The HGP consistency, however, only decreases by 1.2\%. This is due to the inherent ability of Gaussian processes to inflate uncertainty when test samples do not closely align with the training data.

\begin{table*}
    \caption{Correction performance for HGP and TCNN for 7 day propagation with time generalization}
    \renewcommand{\arraystretch}{1.3}
    \centering
    \begin{tabularx}{0.7\textwidth}{|C|C|C|C|C|C|C|}
        \hline
         {Dataset} & $\sigma_{R_{s}}$ km &  $\sigma_{V_{r}}$ m/s & $\sigma_{\|R\|}$ km & $\sigma_{\|V\|}$ m/s &\%{$D_{M_u,C}$} \\
         \hline\hline
         Uncorrected & 12&	13& 11& 12& -\\
         \hline
         TCNN & 11	&12	&11& 12&	64\\
         \hline
         HGP & 9.9 &	11 &	9.3&	10&	96\\
         \hline
    \end{tabularx}
    \label{tab:time-generalization}
\end{table*}

\section{Discussion}
The experimental results for HGP and TCNN corrections show that both models are capable of improving the orbit and covariance accuracy. The prediction accuracy of both models is very similar. They are within 1 km of each other for position and 1 m/s for velocity. The variance prediction accuracy metric is also within 1\% for the two models. While both models reduce the position and velocity error standard deviation by about 50\%, there are still areas of improvement.

Given the HGP model achieved similar performance to the TCNN using only 0.1\% of the training dataset and less than half the feature variables, we likely could reduce the feature set used in the TCNN and increase the size of the training dataset to enable more optimization epochs without overfitting. Furthermore, gradient descent may not find optimal mean predictions when using heteroscedastic Gaussian loss due to under weighting of points that are difficult to model \cite{seitzer2022pitfallsheteroscedasticuncertaintyestimation}. Seitzer et al.  \cite{seitzer2022pitfallsheteroscedasticuncertaintyestimation} suggest using a scale factor in the loss function to adjust the dependence of the gradients on predictive variance. This method could help improve the mean prediction accuracy while still optimizing the variance function.

The HGP model suffered from computational complexity of the large training dataset. Sparse HGPs, such as the one used in \cite{peng2023medium}, optimize a subset of feature vectors that best approximate the larger dataset. This method allows for more data to be used to train the model without scaling up the computational complexity. The challenge with this method is that the training process is more complex since both the primary feature vectors and the kernel parameters must be optimized simultaneously. Given the promising performance of the HGP using a naive downsampling of the full dataset, a sparse method trained on the full dataset could perhaps improve its orbit prediction accuracy.

The covariance corrections from both models generated Mahalanobis distance distributions that were closer to the theoretical $\chi^2$ distribution than the uncorrected distribution. As discussed in section \ref{sec:dataset_generation}, the VCM dataset only contains valid position standard deviations. The true covariance matrix for the VCM initial conditions likely contains correlations and velocity sigmas that would be necessary for an accurate representation of the initial uncertainty. Ultimately, we show promising improvements to covariance realism with our approach but should validate the approach with full covariance information before integration into a navigation filter.

For practical integration of our approach into a navigation pipeline, the machine learning models need to generalize well to new satellites and future VCM epochs. The models trained in this work do not generalize over new VCM epochs most likely due to the limited window of time the training set covers. Depending on the tracking frequency for each satellite the training dataset only spans 1-2 weeks. The training set is therefore unlikely to capture temporal changes to orbit propagation accuracy leading to significant epistemic uncertainty. This is most likely the reason the TCNN was overconfident in its variance predictions. Ensembling approaches could be taken in future work to account for epistemic uncertainty in the TCNN \cite{acharya2023}. Another challenge with generalization across time in LEO orbit propagation is the changes in space weather due to the solar cycle. The solar cycle fluctuates every 11-years \cite{noaa-sun-cycle} and higher solar activity leads to larger errors in atmospheric density modeling \cite{brandt2020simple}. It is impractical to maintain a dataset spanning the full solar cycle, however, future work will investigate if training on a one or two month window improves time generalization for VCM epochs immediately following the training data.

Despite our improvements to orbit accuracy, we recognize that the corrected errors remain on the order of kilometers in the along-track position and meters per second in the radial velocity dimensions. Without further improvements to the models, these errors alone are too large to meet typical PNT system requirements. The remaining dimensions, however, have significantly better propagation performance and remain useful for navigation after a seven day propagation. With enough geometric variation, the remaining four dimensions can provide accurate information for navigation. When a filter ingests the propagated state information, it depends on the covariance to determine which dimensions are accurate. If the propagated covariance is over confident and the true error distribution is not Gaussian, the filter may fail to leverage this accurate navigation information. The main benefit of our current correction performance is therefore improving the Gaussianity of the true error distribution and the covariance realism such that a navigation filter can safely ingest the propagated ephemeris. 

Finally, another practical application of the current modeling performance is to decide which satellites to use for navigation after long propagation times. Given the trained model and at least two VCMs for a newly launched LEO satellite, maintainers of alternative PNT systems can predict the propagation accuracy. Given their mission and system requirements, they can add or reject the satellite from their catalog without additional analysis. To improve on this use case, future work will look at training models without reverse propagation errors as features. Therefore, only information from a single VCM epoch would be required to predict orbit propagation accuracy.

\section{Conclusions}
We have explored using parametric and non-parametric machine learning models to improve orbit and covariance accuracy for different LEO satellite types and orbits. By focusing on LEO, most of the propagation error is due to atmospheric drag mismodeling and therefore accumulates in the argument of latitude. Instead of training the model on the complex error distributions in the Cartesian frame, we perform regression with the one dimensional errors in the argument of latitude.  The argument of latitude prediction from both models reduces the orbit position and velocity accuracy by approximately 50\% for validation satellites disjoint from the training set. The TCNN produced more accurate orbit corrections than the HGP on validation data within the same time window as the training data. However, the HGP covariance predictions generalized better across time. Our method demonstrates the ability of a single model to improve orbit propagation for different LEO satellites. By improving the accuracy of orbit propagation, we extend the time window in which alternative PNT systems can leverage VCMs for navigation.





\acknowledgements 
This work was supported by The Charles Stark Draper Laboratory.

\bibliographystyle{IEEEtran}
\bibliography{research}

\thebiography
\begin{biographywithpic}
{Alex Moody}{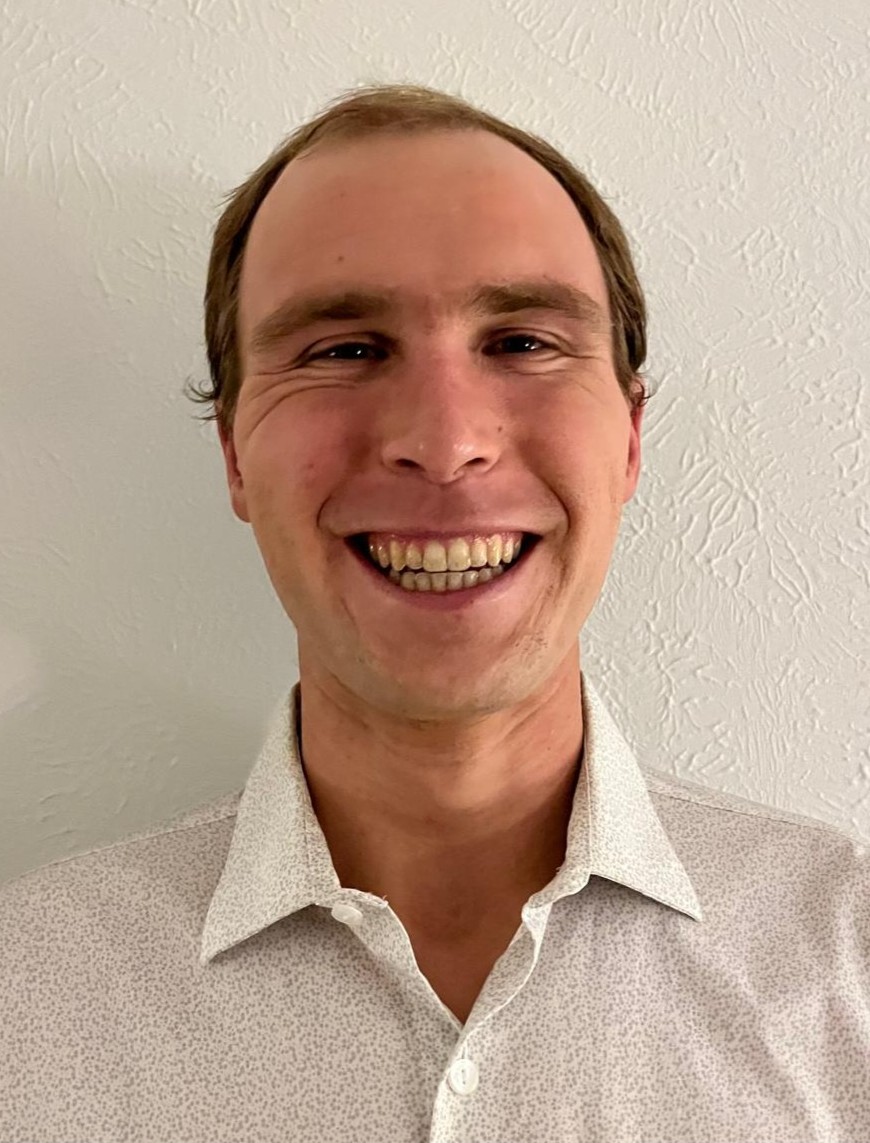}
received his B.S. in Engineering from Harvey Mudd College and is currently a graduate student researcher in the Colorado Center for Astrodynamics Research at University of Colorado, Boulder. His research is funded by Draper and focuses on navigation utility of LEO satellites. Prior to graduate studies, he spent four years developing navigation systems at Draper.
\end{biographywithpic} 

\begin{biographywithpic}
{Penina Axelrad}{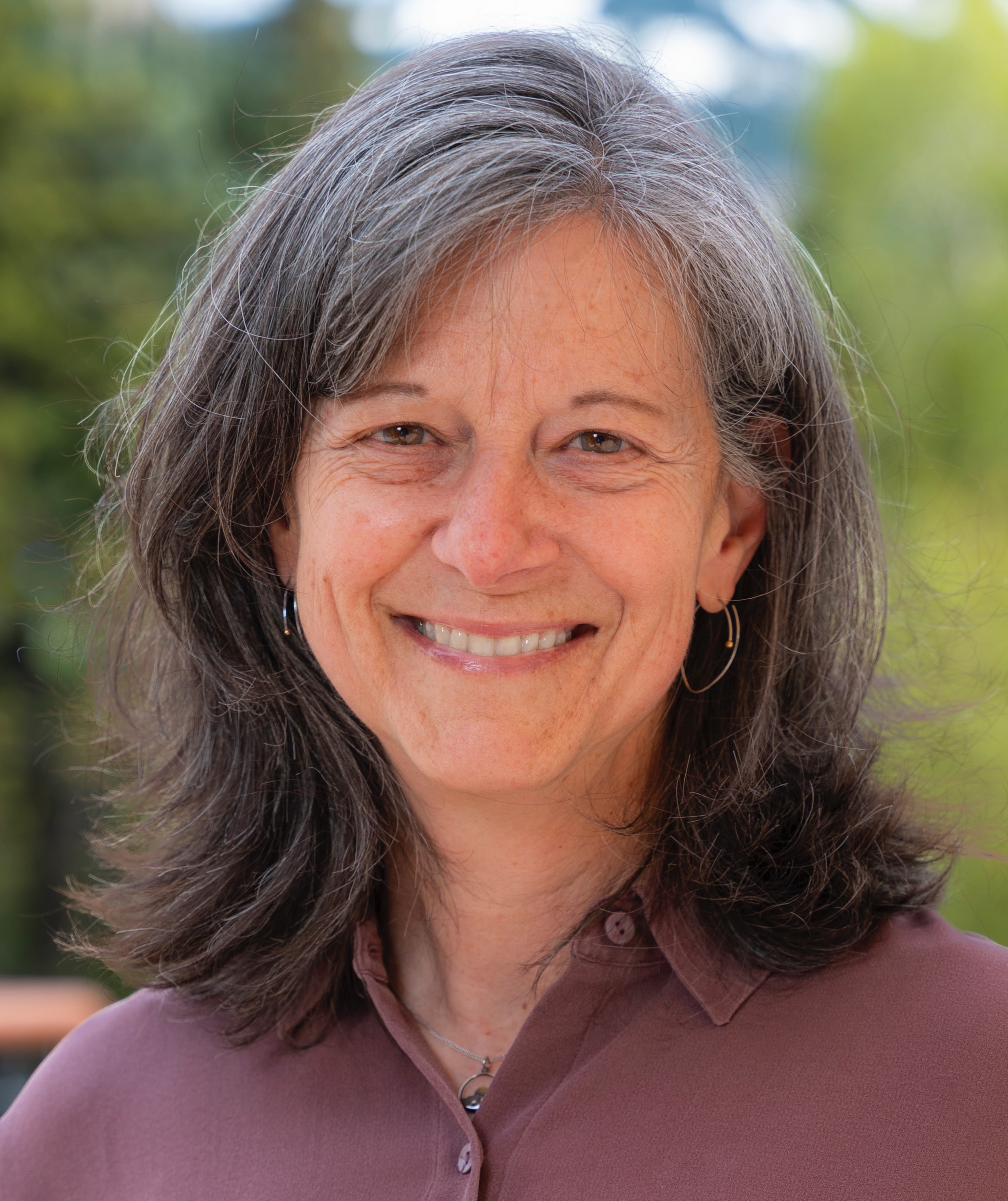}
is a Distinguished Professor in Aerospace Engineering Sciences at the University of Colorado Boulder.  Her research interests include GNSS-based and alternative positioning, navigation, and timing techniques.  Dr. Axelrad is a Fellow of the Institute of Navigation and the AIAA, and a Member of the National Academy of Engineering.

\end{biographywithpic}

\begin{biographywithpic}
{Rebecca Russell}{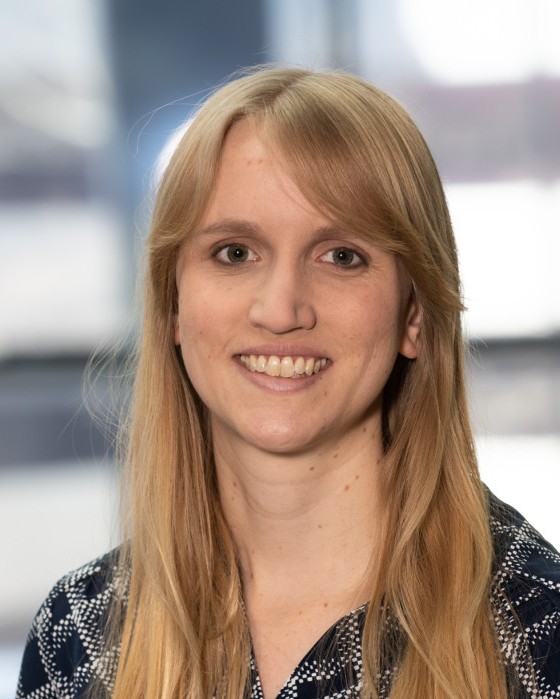}
is a Principal Member of the Technical Staff in the Perception and Embedded Machine Learning Group at Draper. She joined Draper in 2016 after receiving her Ph.D. from MIT and B.S. from Caltech. At Draper, Dr. Russell’s research focuses on using uncertainty-aware deep learning to solve challenges in autonomy and navigation.
\end{biographywithpic}

\end{document}